\documentclass[10pt,twocolumn,letterpaper]{article}

\usepackage{iccv}
\usepackage{times}
\usepackage{epsfig}
\usepackage{graphicx}
\usepackage{amsmath}
\usepackage{amssymb}
\usepackage[pagebackref=true,breaklinks=true,letterpaper=true,colorlinks,bookmarks=false]{hyperref}

\iccvfinalcopy % *** Uncomment this line for the final submission
\def\iccvPaperID{3102} % *** Enter the ICCV Paper ID here

\ificcvfinal\fi
\begin{document}

%%%%%%%%% TITLE
\title{Deep Contextual Attention for Human-Object Interaction Detection}
% \and
% work was done during an internship at IIAI.
% work done at IIAI during Tiancai's internship
\author{Tiancai Wang$^{1}$\thanks{Equal contribution} \thanks{Work done at IIAI during Tiancai's internship.}, Rao Muhammad Anwer$^{2*}$, Muhammad Haris Khan$^{2}$, Fahad Shahbaz Khan$^{2}$,\\ Yanwei Pang$^{1}$, Ling Shao$^2$, Jorma Laaksonen$^3$\\
$^1$School of  Electrical and Information Engineering, Tianjin University\\
% $^2$Inception Institute of Artificial Intelligence, Abu Dhabi, UAE\\
$^2$Inception Institute of Artificial Intelligence (IIAI), UAE\\
$^3$Department of Computer Science, Aalto University School of Science, Finland\\
%$^4$Department of Computer Science, Aalto University School of Science, Finland \\
% {\tt\small pyw@tju.edu.cn, wangtc@tju.edu.cn, rao.anwer@aalto.fi,fahad.khan@liu.se,ling.shao@ieee.org}
{\tt\footnotesize $^1${\{wangtc, pyw\}}@tju.edu.cn,~$^2${\{rao.anwer,~muhammad.haris,~fahad.khan,~ling.shao\}}@inceptioniai.org
% $^3${\{jorma.laaksonen\}}@aalto.fi}
}\\
{\tt\small $^3${\{jorma.laaksonen\}}@aalto.fi}
}
\maketitle
\ificcvfinal\thispagestyle{empty}\fi

%%%%%%%%% ABSTRACT
\begin{abstract}
Human-object interaction detection is an important and relatively new class of visual relationship detection tasks, essential for deeper scene understanding. Most existing approaches decompose the problem into object localization and interaction recognition. Despite showing progress, these approaches only rely on the appearances of humans and objects and overlook the available context information, crucial for capturing subtle interactions between them. We propose a contextual attention framework for human-object interaction detection. Our approach leverages context by learning contextually-aware appearance features for human and object instances. The proposed attention module then adaptively selects relevant instance-centric context information to highlight image regions likely to contain human-object interactions. Experiments are performed on three benchmarks: V-COCO, HICO-DET and HCVRD. Our approach outperforms the state-of-the-art on all datasets. On the V-COCO dataset, our method achieves a relative gain of 4.4\% in terms of role mean average precision ({mAP$_{role} $}), compared to the existing best approach.

\vspace{-0.3cm}
\end{abstract}

%%%%%%%%% BODY TEXT
%-------------------------------------------------------------------------
\section{Introduction}
Recent years have witnessed tremendous progress in various instance-level recognition tasks, including object detection and segmentation. These instance-level problems have numerous applications in robotics, autonomous driving and  surveillance. However, such applications demand a deeper knowledge of scene semantics beyond instance-level recognition, such as the inference of visual relationships between object pairs. Detecting human-object interactions (HOI) is a class of visual relationship detection. Given an image, the task is to not only localize a human and an object, but also recognize the interaction between them. Specifically, it boils down to detecting $\langle$\textit{human, action, object}$\rangle$ triplets. The problem is challenging as it focuses on both human-centric interactions with fine-grained actions (\ie, riding a horse vs. feeding a horse) and involves multiple co-occurring actions (\ie, eating a donut and interacting with a computer while sitting on a chair).

% Recent years have witnessed tremendous progress in various instance-level recognition tasks, including object detection and segmentation. These instance-level problems have numerous applications in robotics, autonomous driving and  surveillance. However, such applications demand a deeper knowledge of scene semantics beyond instance-level recognition, such as the inference of visual relationships between object pairs. Detecting and recognizing human-object interactions (HOI) is a class of visual relationship detection and is essential for deeper understanding of scene semantics. Given an image, the task is to not only localize a human and an object but also recognize the interaction between them. Specifically, it boils down to detecting $\langle$\textit{human, action, object}$\rangle$ triplets. The problem is challenging as it focuses on both human-centric interactions with fine-grained actions (\ie, riding a horse vs. feeding a horse) and involves multiple co-occurring actions (\ie, eating a donut and interacting with computer while sitting on a chair).

\begin{figure*}[t]
\begin{center}
    \includegraphics[width=\linewidth]{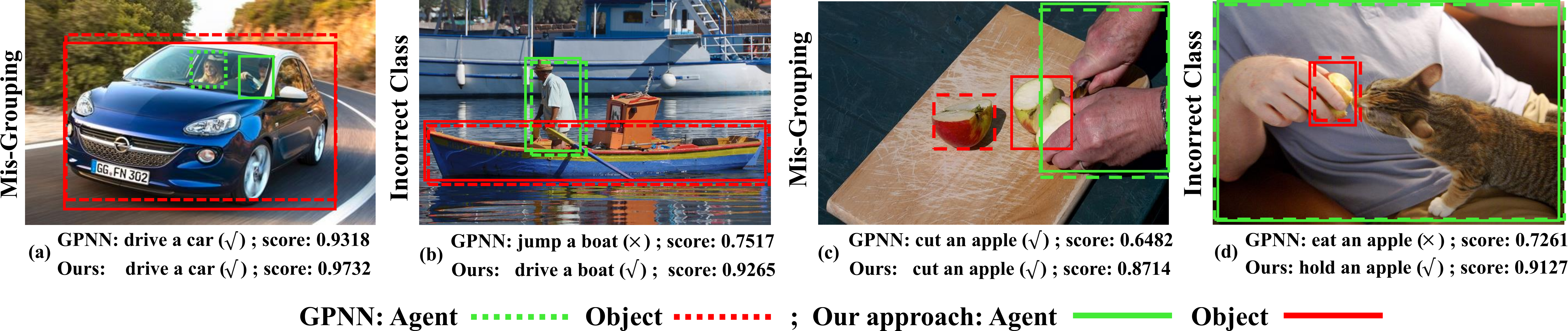}
   %\resizebox{0.8\textwidth}{!}{\includegraphics{S3D.png}}%
\end{center} \vspace{-0.3cm}
   \caption{Example of HOI detections using the proposed approach and the recently introduced GPNN method \cite{qi2018learning}. The four examples depict two HOI detection cases. First in (a) and (b), different object categories (\textit{car} and \textit{boat}) involve the same human-object interaction (\textit{drive}). Second in (c) and (d), different human-object interactions (\textit{cut an apple} and \textit{hold an apple}) involve the same object (\textit{apple}). In case of (a) and (c), GPNN method fails to correctly pair the agent (person) and object, while it miss-classifies the action categories (b) and (d). Our approach accurately groups the agent and the respective object, while correctly classifying the action labels (scores) in all four cases.
   } \vspace{-0.3cm}
   \label{fig:intro_figure}

\end{figure*}

% can be resolved by exploiting information in the local vicinity of a human-object pair (face versus knife).
% \caption{Example HOI detections using the proposed approach and the recently introduced GPNN method \cite{qi2018learning}. There are four different examples depicting two HOI detection cases. First in (a) and (b), context provides additional cue to distinguish different categories involving same interaction (drive). Second in (c) and (d), different interactions involving the same object (apple) can be resolved by exploiting information in the local vicinity of a human-object pair (face versus knife).
% the driving interaction is difficult to distinguish based only on appearance, however, the global surrounding (water and road) could prove helpful. Second in (c) and (d) the interaction on same object is difficult to recognize for the same reason, however, the local surrounding specific to detected proposal is a useful cue.

Most existing HOI detection approaches typically tackle the problem by decomposing it into two parts: object localization and interaction recognition  \cite{chao2018,gkioxari2017interactnet,gupta2015visual,bar2018,qi2018learning,Xu18}. In the first part, off-the-shelf two-stage object detectors \cite{Girshick15ICCV,fasterrcnn_2015_nips,girshick14CVPR} localize both human and object instances in an image. In the second part, detected human and object instances and the pairwise interaction between them are treated separately in a multi-stream network architecture. Recent works have attempted to improve HOI detection by integrating, \eg, structural information \cite{qi2018learning}, gaze and pose cues \cite{Xu18}. Despite these recent advances, the HOI detection performance is still far from satisfactory compared to other vision tasks, such as object detection and instance segmentation.

Current HOI detection approaches tend to focus on appearance features of human and object instances (bounding-boxes) that are central to scoring human-object interactions, and thereby identifying triplets. However, the readily available auxiliary information, such as context, at various levels of image granularity is overlooked. Context information is known to play a crucial role in improving the performance of several computer vision tasks \cite{ding2018context, yao2010modeling, liu2018structure, chen2017}. However, it is relatively underexplored for the high-level task of HOI detection, where context around each candidate detection is likely to provide complementary information to standard bounding-box appearance features. Global context provides valuable image-level information by determining the presence or absence of a specific object category. For instance, when detecting \textit{driving a boat} interaction category, person, boat and water are likely to co-occur in an image. However for \textit{drive a car} category, interaction (drive) remains the same and only context (water) is changed. Besides global context, information in the immediate vicinity of each human/object instance provides additional cues to distinguish different interactions, \eg, various interactions involving the same object. For instance, the surrounding neighborhood in \textit{eating an apple} category is the face of the person whereas for \textit{cutting an apple} category, it is knife and part of the hand (see Fig.~\ref{fig:intro_figure}). In this work, we leverage the context information to the relatively new problem of HOI detection.

\noindent \textbf{Contributions:} We first introduce a contextually enriched appearance representation for human and object instances. While providing auxiliary information, global context also introduces background noise which hampers interaction recognition performance. We therefore propose an attention module to suppress the background noise, while preserving the relevant contextual information. Our attention module is conditioned to specific instances of humans and objects to highlight the interaction regions, \ie, \textit{kick a sports ball} versus \textit{throw a sports ball} categories. The resulting human/object attention maps are then used to modulate the global features to highlight image regions that are likely to contain a human-object interaction.

We validate our approach on three HOI detection benchmarks: V-COCO~\cite{gupta2015visual}, HICO-DET~\cite{chao2018} and HCVRD ~\cite{HCVRD18}. We perform a thorough ablation study to show the impact of context information for HOI detection. The results clearly demonstrate that the proposed approach provides a significant improvement over its non-contextual baseline counterpart. Further, our contextual attention-based HOI detection framework sets a new state-of-the-art on all datasets. On HICO-DET dataset, our approach yields a relative gain of 9.4\% in terms of mean average precision (mAP), compared to the best published method \cite{gao2018ican}. Fig.~\ref{fig:intro_figure} shows a comparison of our approach with GPNN \cite{qi2018learning} on HICO-DET images.

\begin{figure*}[t]
\begin{center}
    \includegraphics[width=\linewidth]{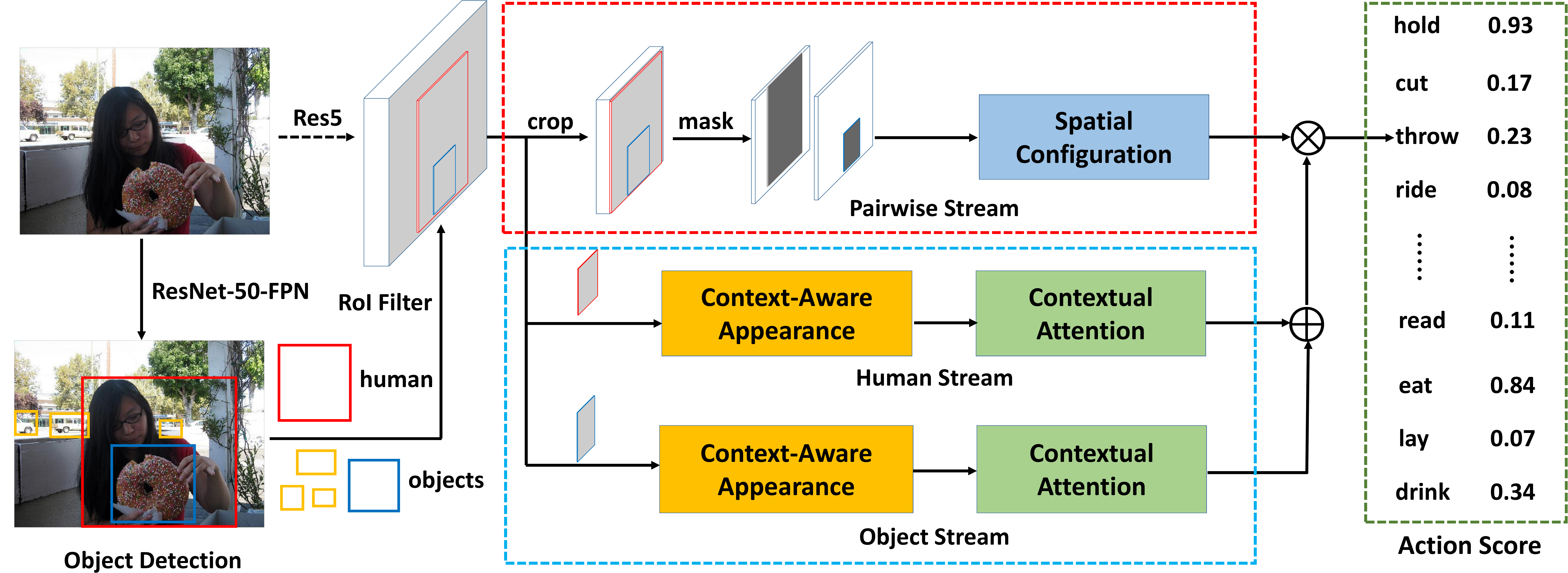}
   %\resizebox{0.8\textwidth}{!}{\includegraphics{S3D.png}}%
\end{center} \vspace{-0.3cm}
   \caption{Overall multi-stream architecture of our proposed HOI detection framework comprising a localization and an interaction stage. For localization, we follow the standard object detector \cite{lin:fpn17} to obtain human and object bounding-box predictions. For interaction prediction, we fuse scores from a human, an object, and a pairwise stream. We introduce context-aware appearance and contextual attention modules in the human and object streams. Final predictions are obtained by fusing the scores from human, object and pairwise streams.} \vspace{-0.2cm}
   \label{fig:overall_architecture}

\end{figure*}

\section{Related Work}
%% object detection

%In recent years, tremendous progress has been made in the field of object detection  \cite{Girshick15ICCV,fasterrcnn_2015_nips,girshick14CVPR,Zhang18,lin:fpn17,JRedmon16CVPR,WLiu16ECCV}, predominantly due to deep convolutional neural networks (CNNs).
\noindent \textbf{Object Detection:} Significant progress has been made in the field of object detection  \cite{Girshick15ICCV,KhanTIP15,fasterrcnn_2015_nips,girshick14CVPR,Zhang18,lin:fpn17,JRedmon16CVPR,WLiu16ECCV}, predominantly due to deep convolutional neural networks (CNNs). Generally, CNN-based object detectors can be divided into two-stage and single-stage approaches. In the two-stage approach, object detection methods \cite{Girshick15ICCV,fasterrcnn_2015_nips,girshick14CVPR} first employ an object proposal generator to generate regions of interests, which are then passed through an object classification and bounding-box regression pipeline. In contrast, single-stage detection methods \cite{JRedmon16CVPR,WLiu16ECCV} directly learn object category predictions (classification) and bounding-box locations (regression) using anchors to predict the offsets of boxes instead of coordinates. Two-stage object detectors are generally more accurate compared to their single-stage counterparts. As in previous HOI detection works \cite{gkioxari2017interactnet, chao2018}, we employ an off-the-shelf two-stage FPN detector \cite{lin:fpn17} to detect both human and object instances.\\
\noindent \textbf{Human-Object Interaction Detection:} Gupta and Malik \cite{gupta2015visual} were the first to introduce the problem of visual semantic role labeling. In this problem, the aim is to detect a human, an object, and label the interaction between them. Gkioxrari \etal, \cite{gkioxari2017interactnet} proposed a human-centric approach by extending the Faster R-CNN pipeline \cite{fasterrcnn_2015_nips} with an additional branch to classify both actions and action-specific probability density estimation over the target object location. The work of \cite{qi2018learning} proposed a Graph Parsing Neural Network (GPNN) in which HOI structures are represented with graphs and then optimal graph structures are parsed in an end-to-end fashion. The work of \cite{Xu18} introduced a human intention-driven approach, where both pose and gaze information are exploited in a three-branch framework: object detection, human-object pairwise interaction and gaze-driven stream. Kolesnikov \etal, \cite{bar2018} proposed a joint probabilistic model for detecting visual relationships. Chao \etal, \cite{chao2018} introduced a human-object region-based CNN approach that extends the region-based object detector (Fast R-CNN) and has three streams: human, object and pairwise. Further, they introduced a new large-scale human-object interaction detection benchmark (HICO-DET). \\
%The proposed approach adopts a multi-stream framework, similar to \cite{chao2018}, as a building block.
\noindent \textbf{Contextual Cues in Vision:} Context provides an auxiliary cue for several vision problems, such as object detection \cite{liu2018structure,chen2017}, action recognition \cite{yao2010modeling}, and semantic segmentation \cite{ding2018context}. Recently, learnable context has gained popularity with the advent of deep neural networks \cite{Girdhar17,liu2018structure}. Despite its success in several tasks \cite{peng2017large, Girdhar17,liu2018structure, BohanZhuang17context,JiananLi17, JiahuiYu18}, the impact of contextual information to the relatively new task of HOI detection is yet to be fully explored.

\section{Overall Framework}\label{subsection:baseline_arch}
The overall framework comprises two stages: localization and interaction prediction (see Fig.~\ref{fig:overall_architecture}). For localization, we follow the popular paradigm of FPN \cite{lin:fpn17} as a standard object detector to generate bounding-boxes for all possible human and object instances in the input image. For interaction prediction, following \cite{chao2018}, we fuse scores from the three individual streams: a human, an object, and a pairwise. Scores from human and object streams
are added. The resulting scores are then multiplied with
pairwise stream. \\
% We now briefly describe the multi-stream pipeline.
\noindent \textbf{Multi-Stream Pipeline:} The inputs to the multi-stream architecture are the bounding-box predictions from FPN  \cite{lin:fpn17} and the original image. The output of the multi-stream architecture is a detected $\langle$\textit{human, action, object}$\rangle$ triplet. The overall framework comprises three separate streams: human, object and pairwise interaction. Both the human and object streams are appearance oriented; they employ CNN feature extraction to generate confidence scores on the detected human and object bounding-boxes. The pairwise interaction stream encodes the spatial relationship between the person
and object as in \cite{chao2018}.
% the next line
% In this work, we use ResNet \cite{resnet16} as our basic feature extractor. However, our approach is generic and can be employed with other backbone architectures.

\begin{figure*}[t]
\begin{center}
\includegraphics[width=\linewidth]{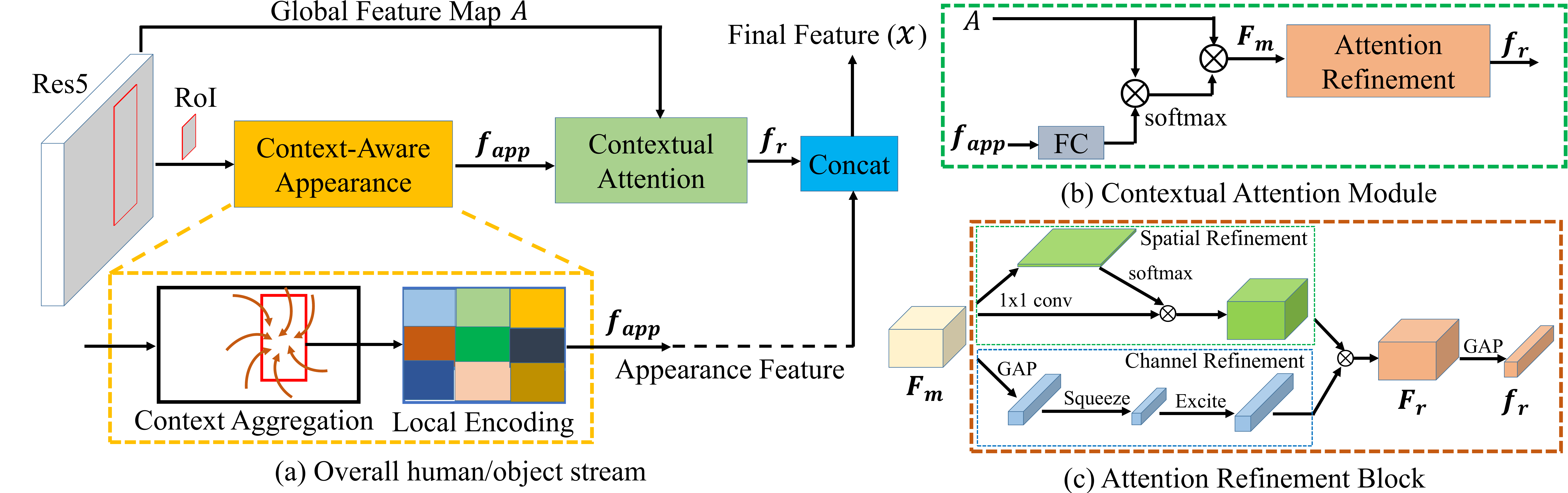}
%\resizebox{0.8\textwidth}{!}{\includegraphics{S3D.png}}%
\end{center} \vspace{-0.3cm}
   \caption{On the left (a), the proposed overall human/object stream. Both the contextual attention module (b) and the attention refinement block (c) are shown on the right. The context-aware appearance module produces contextual appearance features that encode both appearance and context information. The contextual appearance features are then fed into the contextual attention module to suppress the background noise resulting in a modulated feature representation. The modulated feature representation is further enriched in the attention refinement block to obtain refined modulated features. Consequently, both contextual appearance and refined modulated features are concatenated to obtain action predictions from the human/object stream.  } \vspace{-0.2cm}
%   Visual description of how the proposed context-aware appearance is generated and the designed deep contextual attention mechanism is used to produce modulated features. To obtain context-aware appearance, the Res5 features follow a context aggregation as explained in text. To obtain modulated features, we produce a deep contextual attention map, relying on these context-aware features, and use it to re-weight the Res5 features.} \vspace{-0.2cm}
   \label{fig:context_attention}
\end{figure*}

\subsection{Proposed Human/Object Stream}
%\label{subsection:hu}am-objnce-stretma

The standard multi-stream architecture encodes instance-centric (bounding-box) appearance features in the human and object streams and ignores the associated contextual information. In this work, we argue that the bounding-box appearance alone is insufficient and that the contextual information in the vicinity of a human and object instances provides complementary information useful to distinguish complex human-object interactions. We therefore enrich the human and object streams (see Fig.~\ref{fig:context_attention}) with contextual information by introducing contextually-aware appearance features $f_{app}$ (sec.~\ref{subsection:context_features}). These contextual appearance features $f_{app}$ are then fed into the contextual
attention module (sec.~\ref{subsection:context_attention}), where they are used to modulate the global feature map $A$ to obtain a modulated feature representation $F_{m}$. The modulated feature representation $F_{m}$ is further refined in the attention refinement block to obtain the refined modulated features $F_{r}$, which further passes through global average pooling to obtain refined modulated vector $f_{r}$. Subsequently, both representations $f_{app}$ and $f_{r}$ are concatenated to obtain action predictions from the human/object streams. Note that the same architecture is employed for both the human and object streams. Thus, the only difference between the two streams is their inputs, which are human and object bounding-box predictions, respectively. Next, we describe different components of our proposed human/object stream.

\vspace{-0.3cm}
\subsubsection{Contextually-Aware Appearance Features}
\label{subsection:context_features}
% motivation:  context is known to be very usefull for improving performance on detection and segmentions tasks using deep learning. serve as to be an effective global contextual prior. size of receptive fields determines how much we use context information. global average pooling good baseline as the global contextual prior also commonly used in image classification task but not enough to cover necessary information .Context information has played a crucial role in improving the performance of different vision tasks such as object recognition \cite{},  object detection \cite{}, and action recognition \cite{}.
%Context is an important aspect in computer vision and has shown to improve performances in several tasks such as, object detection \cite{liu2018structure,chen2017}, semantic segmentation \cite{ding2018context}, and action recognition \cite{yao2010modeling}.

%The human and object streams within the multi-stream architecture, explained above, encode \textit{only} instance-centric appearance features from the human and object bounding-box, respectively.

Given the CNN features (Res5 block of the ResNet-50 backbone) of the whole image, as well as human/object bounding-box predictions from the detector, standard instance-centric appearance features are extracted by employing region-of-interest (ROI) pooling followed by a residual block and global average pooling. Though theoretically the image-level CNN features used in the construction of the standard appearance representation are supposed to cover entire spatial image extent, their valid receptive field is much smaller in practice~\cite{zhou14}. This implies that the larger global scene context prior is ignored in such a standard appearance feature construction. Our context-aware appearance module is designed to capture additional context information and consists of context aggregation and local encoding blocks (see Fig.~\ref{fig:context_attention}(a)).
%The context aggregation block operates on the CNN features of the whole image whereas the local encoding block aims at improving the ROI pooled features.

The context aggregation block aims to capture a larger field-of-view (FOV) to integrate context information in instance-centric appearance features, while preserving spatial information. A straightforward way to capture a larger FOV is through a fully connected (FC) layer or cascaded dilated convolutions. However, the former collapses spatial dimensions, while the latter produces sparser features. Therefore, our context aggregation block employs a large convolutional kernel (LK) previously used for semantic segmentation \cite{peng2017large}. To the best of our knowledge, we are the first to introduce a large kernel-based context aggregation block to construct contextual appearance features for the problem of HOI detection. The input to the context aggregation block is the CNN features (Res5 block) of the image with size $h\times w\times c{_{in}}$, where $c_{in}$ denotes the number of channels and $h$ and $w$ denote the input feature dimensions. The output of the context aggregation block is then context-enriched features of size $h\times w\times c{_{out}}$, obtained after applying a large kernel of size $k\times k$ to the original CNN features. In this work, we utilize the factorized large kernel, which is efficient as its computational complexity and number of parameters are only $O(2/k)$, compared to the trivial $k\times k$ convolution.

Beside context aggregation, our context-aware appearance module contains a local encoding block. Existing HOI detection approaches employ standard ROI warping, which involves a max-pooling operation performed on the cropped ROI region.
Our local encoding block aims to preserve locality-sensitive information in each bounding-box ROI region by encoding the position information with respect to a relative spatial position. Such a strategy has been previously investigated to encode spatial information within ROI regions in the context of generic object detection \cite{dai2016r}. However, \cite{dai2016r} directly employs a $1\times 1$ convolution on the standard CNN feature map (Res5). Instead, we encode locality-sensitive information in each ROI region based on the contextualized CNN feature map obtained from our context aggregation block. Further, \cite{dai2016r} utilizes PSRoIpooling with average pooling. Instead, we employ the PSRoIAlign together with max-pooling. PSRoIAlign is employed to reduce the impact of coarse quantization caused by PSRoIpooling through bilinear interpolation. Fig.~\ref{fig:psp} shows the impact of PSRoIAlign-based local encoding on the input feature maps of an image. Consequently, the output of the local encoding block is flattened and passed through a fully-connected layer to obtain conextual appearance features $f_{app}$.

\vspace{-0.3cm}
\subsubsection{Contextual Attention}
\label{subsection:context_attention}
The contextual appearance features, described above, encode both appearance and global context information. However, not all background information is equally useful for the HOI problem. Further, integrating meaningless background noise can even deteriorate the HOI detection performance. Therefore, a careful identification of useful contextual information is desired to distinguish subtle human-object interactions that are difficult to handle otherwise. Generally, attention mechanisms are used to highlight the discriminative features particularly important for a given task \cite{Tsotsosattention}. The contextual attention module in our human/object stream consists of bottom-up attention and attention refinement components. The bottom-up attention component is based on the recently introduced approach of \cite{Girdhar17} for action recognition and exploits a scene-level prior to focus on relevant features. Note, \cite{Girdhar17} computes image-level attention, whereas we aim to generate bounding-box based attention. Further, contrary to standard appearance features, the bottom-up attention maps in our attention module are generated using \textit{contextually-aware appearance} features $f_{app}$ (sec.~\ref{subsection:context_features}) that encode both appearance and context. We generate modulated features by first constructing a contextual attention map, which is then deployed to modulate the input CNN feature map (see Fig.~\ref{fig:context_attention}(b)).

Specifically, we project the input (Res5) feature maps $f$ using a 1$\times$1 convolution onto a 512-dimensional space, denoted as $A$. Then, we compute the dot product between these projected global features $A$ and contextual-appearance features  $f_{app}$ to obtain an attention map, which is then used to modulate $A$, such that,

 \begin{equation}
  %z = softmax((G{_{context}(f),B}) \otimes A) \otimes A,\label{Eq:mod_features}
  F_{m} = \text{softmax}(f_{app} \otimes A) \otimes A   \label{Eq:mod_features}
  \end{equation}

Here, $F_{m}$ are the resulting modulated features. The discriminative ability of $F_{m}$ is further enhanced in the attention refinement block, which consists of spatial and channel-wise attention refinement. The attention refinement block is simple and light-weight (see Fig.~\ref{fig:context_attention}(c)). During spatial refinement, we first apply a 1$\times$1 conv on modulated features $F_{m}$ to generate a single-channel heatmap $H$, followed by a softmax-operation-based normalization. Then, we perform an element-wise multiplication between the normalized heatmap and the modulated features $F_{m}$.  The resulting spatial refinement $S_{att}$ learns the most relevant features as:

\begin{figure}[t]
\begin{center}
    \includegraphics[width=\linewidth]{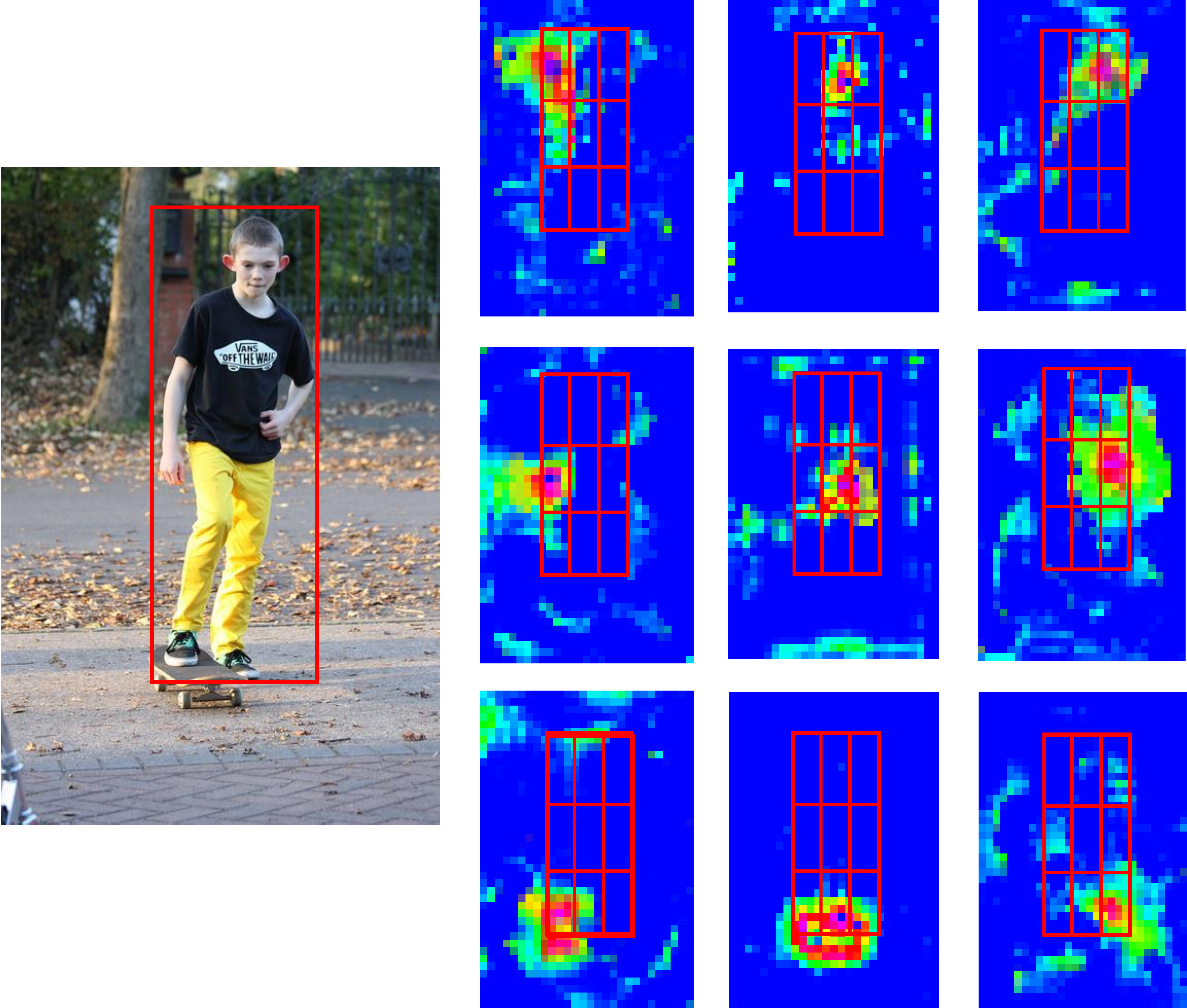}
   %\resizebox{0.8\textwidth}{!}{\includegraphics{S3D.png}}%
\end{center} \vspace{-0.3cm}
   \caption{Visual depiction of the local encoding block that preserves locality-sensitive information. For illustration purposes, the detected human bounding-box is divided into 3$\times$3 sub-regions and there are 9 score maps. Each sub-region votes for the presence of a specific object part, relative to the position of the object, based on how good the bounding-box overlaps with the score maps. } \vspace{-0.3cm}
   \label{fig:psp}

\end{figure}
%\subsubsection{Hybrid Attention Module}
% We further boost the discriminative ability of the modulated feature by proposing a hybrid attention module. The proposed module is simple and light-weight as shown in Fig. \ref{fig:hybrid_module}. It comprises two components: spatial attention and channel-wise attention.% and formulated as:
% \begin{equation}
%  z_{hybrid} = S_{att}(z) \otimes C_{att}(z),
%  \label{Eq:hybrid}
% \end{equation}
% \indent where $S_{att}$ and $C_{att}$ are the spatial and channel-wise attention operations, and $z_{hybrid}$ is the modulated feature after hybrid attention module.
%We propagate the above generated attention maps from contextual attention module to further enhance the instance-guided context feature.  We propose a hybrid attention module as a next attention step which consists of two components: spatial attention and channel-wise attention.
%\noindent \textbf{Spatial Attention:}

% The attention refinement block is simple and light-weight as shown in Fig. \ref{fig:hybrid_module}.

% We formulate \textbf{spatial attention} operation, as shown in green dashed box in Fig. \ref{fig:hybrid_module}, by first applying 1x1 conv on $f_{m}$ to generate a single channel heatmap $H$ and normalize them with a softmax operation. Next, we perform an element-wise product between the normalized heatmap and $f_{m}$ feature. The spatial attention $S_{att}$ enforces to learn the most relevant region of the feature and is described as:
%\begin{equation}
%H = softmax((a(z)^T)
%\end{equation}

\begin{equation}
S_{att}(F_{m}) = H \otimes F_{m}  \label{Eq:spatial_att}
\end{equation}
%\noindent \textbf{Channel-wise Attention:}

Beside spatial refinement, we also perform a channel-wise refinement. Inspired by the squeeze-and-excitation network (SENet) of \cite{hu2017squeeze}, we first apply global average pooling on the modulated features $F_{m}$ to squeeze global spatial information into a channel descriptor $z$. Then, the excitation stage is a stack of two FC layers, followed by a sigmoid activation with input $z$ and is described as:

\begin{equation}
C_{att}(F_{m}) = \sigma (W_{1}\delta (W_{2}z))
\label{Eq:channel_att}
\end{equation}

Here, $z$ is the output of the squeeze operation, and $W_{1}$ and $W_{2}$ refer to fully-connected operations. $\delta$ and $\sigma$ are ReLU and sigmoid activations, respectively. Finally, $C_{att}$ modulates the spatially-attended features $S_{att}$ to further highlight regions relevant to human-object interaction to obtain a refined modulated feature representation $F_{r}$ as:

\begin{equation}
 F_{r} = S_{att}(F_{m}) \otimes C_{att}(F_{m})
 \label{Eq:hybrid}
\end{equation}
%Eq. \ref{Eq:hybrid}.

% \indent We aim to exploit the contextual information contained in the feature itself through \textbf{channel-wise attention}. We follow the squeeze-and-excitation network (SENet) style \cite{hu2017squeeze} as shown in blue dashed box in Fig. \ref{fig:hybrid_module}. Global average pooling (GAP) is first applied on $f_{m}$ to squeeze global spatial information into a channel descriptor $z^{'}$. Then the excitation stage is a stack of two fully-connected layers followed by sigmoid activation
% with input $z^{'}$. We formulate channel-wise attention as:

% \begin{equation}
% C_{att}(f_{m}) = \sigma (W_{1}\delta (W_{2}z^{'})),
% \label{Eq:channel_att}
% \end{equation}

% \indent Where $z^{'}$ is the output of squeeze operation, $W_{1}$ and $W_{2}$ refer to fully-connected operation. $\delta$ and $\sigma$ are ReLU and sigmoid activations, respectively. Finally, $C_{att}$ modulates the spatially-attended feature $S_{att}$ to further highlight regions relevant to detection to get the output $f_{r}$ and formulated as:

% %is the modulated feature after hybrid attention module.

% \begin{equation}
%  f_{r} = S_{att}(f_{m}) \otimes C_{att}(f_{m}),
%  \label{Eq:hybrid}
% \end{equation}
% %Eq. \ref{Eq:hybrid}.

%\noindent \textbf{Combining Contextual and Hybrid Attentions:}
% \subsubsection{Combining Contextual Appearance and Attention Features}
% \label{subsection:combined_context_attention}
Finally, the refined modulated features $F_{r}$ are passed through global average pooling to obtain the refined modulated vector $f_{r}$. We combine contextual appearance features $f_{app}$ and the refined modulated vector $f_{r}$ to produce the final representation $x$. This representation $x$ is then passed through two FC layers to estimate action predictions from the human/object stream, respectively. Given an HOI predicted bounding-box, the final prediction is obtained by fusing the scores from the human, object and pairwise streams.

\section{Experiments}
%-------------------------------------------------------------------------
% We first introduce the two standard benchmarks and their evaluation protocols, followed by relevant implementation details of the proposed framework. We then report a thorough baseline comparison. Finally, we show a comparison of our proposed framework with the state-of-the-art methods on both benchmarks.

\subsection{Dataset and Evaluation Protocol}
% We validate the performance of our proposed approach on two popular HOI detection benchmarks: V-COCO \cite{gupta2015visual} and HICO-DET \cite{chao2018}.

\noindent\textbf{V-COCO \cite{gupta2015visual}:} is the first HOI detection benchmark and a subset of popular MS-COCO dataset \cite{coco14}. The V-COCO dataset contains 10,346 images in total, with 16,199 human instances. Each human instance is annotated with 26 binary action labels. Note that three action classes (\textit{i}.\textit{e}., cut, hit, eat) are annotated with two types of targets (\textit{i}.\textit{e}., instrument and direct object). It includes 2533, 2867, and 4946 images for training, validation and testing, respectively.
% We follow the evaluation protocol provided with the dataset \cite{gupta2015visual}.

%We use role mean Average Precision (role mAP) as evaluation protocol for V-COCO following \cite{gkioxari2017interactnet}. The aim is to detect the $\langle$\textit{human, action, object}$\rangle$ triplet. It is considered as true positive if both the predicted human and object boxes has a IoU of 0.5 or higher with the corresponding ground truth boxes and the predicted and ground truth actions match.

\noindent\textbf{HICO-DET \cite{chao2018}:} is a challenging dataset and has 47,776 images in total, with 38,118 images for training and 9658 images for testing. There are more than 150k human instances annotated with 600 types of different human-object interactions. The HICO-DET dataset contains same 80 object categories as MS-COCO and 117 action verbs.

\noindent\textbf{HCVRD \cite{HCVRD18}:} is a large-scale dataset and is labeled with both human-centric visual relationships and corresponding ‘human’ and ‘object’ bounding boxes. It has 52,855 images with 1,824 object categories and 927 predicates. It contains 256,550 relationships instances  and there are on average 10.63 predicates per object category. We evaluate our method on the predicate detection task, where the goal is to perform predicate recognition given the labels and bounding boxes for both object and human.

\noindent \textbf{Evaluation Protocol:}
We use the original evaluation protocols for all three datasets, as provided by their respective authors. For the V-COCO dataset, we use role mean Average Precision ({mAP$_{role} $}) as an evaluation metric. Here, the aim is to detect the $\langle$\textit{human, action, object}$\rangle$ triplet. The HOI detection is considered correct if the intersection-over-union (IoU) between the human and object bounding-box predictions and the respective ground-truth boxes is greater than the threshold 0.5 together with the correct action label prediction. For HICO-DET, results are reported in terms of mean average precision (mAP). For HCVRD, we report
top-1 and top-3 results at 50 and 100 recall.

% In case of HICO-DET, results are reported in terms of mean average precision (mAP).
% It is taken as true positive if both the predicted human and object boxes has an intersection-over-union (IoU) of 0.5 or higher with respect to the corresponding ground truth boxes and the predicted and ground truth actions match. Different to V-COCO, we report the results in mean Average Precision (mAP) as descirbed in \cite{chao2018}. % the standard object detection evaluation

\subsection{Implementation Details}
We deploy Detectron \cite{Detectron2018} with a ResNet-50-FPN \cite{lin:fpn17} backbone to obtain human and object bounding-box predictions. To select a predicted bounding-box as a training sample, we set the confidence threshold to be higher than 0.8 for humans and 0.4 for objects. For interaction prediction, we employ ResNet-50 as the feature extraction backbone pre-trained on ImageNet. The initial learning rate is set to 0.001, weight decay of 0.0001 and a momentum of 0.9 is used for all datasets. The network is trained for 300k on V-COCO and 1800k iterations on HICO-DET and HCVRD, respectively. For input image of size $480 \times 640$, our interaction recognition part of the approach takes 130 milliseconds (ms) to process, compared to its baseline counterpart (111ms) on a Titan
X GPU.

% Well-documented code together with pre-trained models will be released upon publication.

%When selecting the training samples of the bounding box, we use the human bounding boxes with confidence score higher than 0.8 and the object bounding boxes with confidence score higher than 0.4. Our network use the ResNet-50 as the feature extraction backbone, which is pre-trained on the COCO dataset. When training our network on both two datasets, a learning rate of 0.001, a weight decay of 0.0001 and a momentum of 0.9 are used. The only difference is that our network is trained 300k iterations on the V-COCO dataset while 1800k iterations on the HICO-DET dataset.
%%%%%-----------------------------------

%%%%------------------------------

\begin{table}[t]
\begin{center}
\resizebox{\linewidth}{!}{
\begin{tabular}{lcccc}
\hline
Add-on  &Baseline  &\multicolumn{2}{c}{}  \\
\hline
\textit{Res5-share}                 &\checkmark  &\checkmark     &\checkmark\\
\textit{Context-aware appearance} (sec.~\ref{subsection:context_features})        &{}          &\checkmark     &\checkmark\\
\textit{Contextual attention } (sec.~\ref{subsection:context_attention})            &{}          &{}          &\checkmark\\
% \textit{local encoding}             &{}          &{}           &{}         &\checkmark &\checkmark\\
%\textit{Hybrid attention}           &{}          &{}           &{}         &\checkmark\\
\hline
mAP$ _{role} $                               &44.5        &46.0              &\textbf{47.3}\\
\hline
\end{tabular}
}
\end{center}\vspace{-0.2cm}
\caption{A baseline comparison when integrating our proposed context-aware appearance and contextual attention modules into the multi-stream architecture. Results are reported in terms of role mean average precision ({mAP$ _{role}) $} on the V-COCO dataset. For fair comparison, we use the same feature backbone (Res 5 block of ResNet-50) for both our approach and the baseline. Both context-aware appearance and contextual attention modules contribute in the overall improvement in HOI detection performance. Our overall architecture achieves a relative gain of 6.3\% over the baseline. }\vspace{-0.1cm}
\label{tab:slice_experiment}
\end{table}

\begin{table}[t]
\begin{center}
\resizebox{\linewidth}{!}{
\begin{tabular}{l|c c c c c}
\hline
Overlap thresh &0.1 &0.3  &0.5 &0.7 &0.9 \\
\hline
Baseline &50.1 &47.8  &44.5 &35.9 &2.5\\
%\hline
Our Approach  &\textbf{53.5} &\textbf{50.8} &\textbf{47.3} &\textbf{37.0} &\textbf{2.8} \\
\hline
\end{tabular}
}
\end{center}\vspace{-0.2cm}
\caption{Performance (in terms of {mAP$ _{role} $}) with different IoU thresholds, used in the testing, to compare the classification capabilities of our approach with the baseline on the V-COCO dataset. The performance gap between our approach and the baseline increases at lower threshold values.}\vspace{-0.1cm}
\label{tab:over_thresh}
\end{table}

\begin{table}[t]
\begin{center}
\resizebox{\linewidth}{10mm}{
\begin{tabular}{l|c|c}
\hline
Backbone  Architecture    & Baseline   & Our Approach \\
\hline
%\hline
VGG-16& 42.0& \textbf{44.5} \\
ResNet-50& 44.5 & \textbf{47.3} \\
ResNet-101& 45.0& \textbf{47.8} \\

\hline
%\hline
\end{tabular}
}
\end{center}\vspace{-0.2cm}
\caption{A comparison (in terms of {mAP$ _{role} $}) of our approach with the baseline when using different backbone network architectures on the V-COCO dataset. Our approach always provides consistent improvements over the baseline using different backbones.} \vspace{-0.2cm}
\label{tab:v-generalization}
 \end{table}

 \begin{table}[t]
\begin{center}
%\resizebox{\linewidth}{10mm}{
\begin{tabular}{l|c|c}
\hline
Methods     & Feature Backbone   & \textbf{mAP$ _{role} $} \\
\hline
%\hline
Gupta et .al\cite{gupta2015visual}*& ResNet-50-FPN& 31.8 \\
InteractNet  \cite{gkioxari2017interactnet}& ResNet-50-FPN & 40.0 \\
BAR \cite{bar2018}& Inception-ResNet& 41.1 \\
GPNN \cite{qi2018learning}& ResNet-50& 44.0 \\
iCAN \cite{gao2018ican}& ResNet-50& 45.3 \\
\hline
 Our Approach & ResNet-50 & \textbf{47.3} \\
\hline
%\hline
\end{tabular}
%}
\end{center}\vspace{-0.2cm}
\caption{State-of-the-art comparison on the V-COCO dataset. * refers to implementation of the approach of \cite{gupta2015visual} by \cite{gkioxari2017interactnet}. The scores are reported in {mAP$ _{role} $} and the best result is in bold. Our approach sets a new state-of-the-art on this dataset, achieving an absolute gain of 2.0\% over the best existing method.} \vspace{-0.3cm}
\label{tab:v-coco-compare}
 \end{table}

%\subsection{Results on V-COCO}
%We compare our approach with the existing state-of-the-art methods in terms of {mAP$ _{role} $} on V-COCO in Table \ref{tab:v-coco-compare}. The proposed approach sets a new state-of-the-art in HOI detection by improving the previous best performing method with a significant difference of 1.7 points. Figure \ref{fig:V-COCO-camparsion} shows some qualitative results produced by our model.

\subsection{Results on V-COCO Dataset}
\noindent \textbf{Baseline Comparison:} We first evaluate the impact of integrating our proposed context-aware appearance (sec.~\ref{subsection:context_features}) and contextual attention (sec.~\ref{subsection:context_attention}) modules into the human/object stream of the multi-stream architecture. Tab.~\ref{tab:slice_experiment} shows the results on the V-COCO dataset. The baseline multi-stream architecture contains standard appearance features from the \textit{Res5} block of the ResNet-50 backbone, which have a size of $h \times w \times 2048$. These standard appearance features are directly passed through the classifier to obtain the final action scores in the human/object stream, achieving a {mAP$ _{role} $} of $44.5$. The introduction of the proposed contextual appearance features improves the HOI detection performance from $44.5$ to $46.0$ in terms of {mAP$ _{role} $}. The performance is further improved by $1.3\%$, in terms of {mAP$ _{role} $} when integrating our proposed contextual attention module. Our final framework achieves an absolute gain of $2.8\%$ in terms of {mAP$ _{role} $}, compared to the baseline.

% that are directly passed through the classifier (without contextual attention module) to obtain the final action scores in human/object stream. These standard appearance features are from \textit{res5} block of size $h \times w \times 2048 $ and achieves a role mAP of $44.5$. The introduction of proposed contextual appearance features improves the HOI detection performance from $44.5$ to $46.5$ in terms of role mAP. The HOI detection performance is further improved by $1.2\%$, in terms of role MAP, when integrating our proposed contextual attention module. Our final framework achieves an absolute gain of $2.7\%$ in terms of role mAP, compared to the baseline.
\begin{figure*}[t]
\begin{center}
    \includegraphics[width=\linewidth]{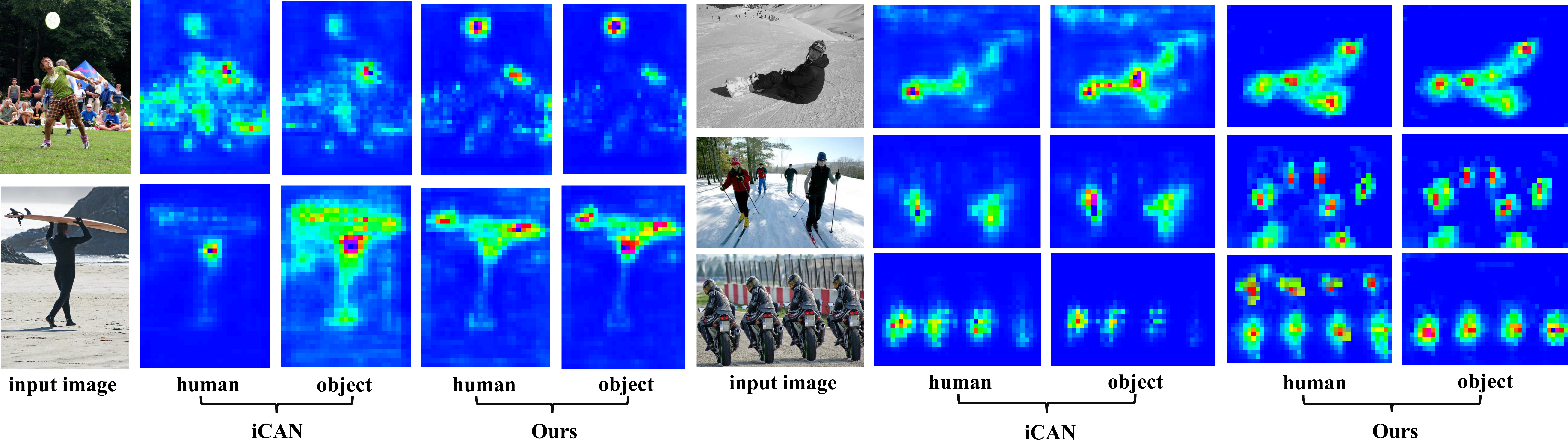}
   %\resizebox{0.8\textwidth}{!}{\includegraphics{S3D.png}}%
\end{center} \vspace{-0.3cm}
   \caption{Comparison of attention maps obtained using our approach and iCAN \cite{gao2018ican} on example images from the V-COCO dataset. Human and object attention maps in iCAN are constructed using standard appearance features. In contrast, human and object attention maps in our approach are constructed using contextual appearance features extracted using the context aggregation and local encoding blocks in our context-aware appearance module. We show examples for both single and multiple human-object interactions.} \vspace{-0.3cm}
   \label{fig:compare_attention}
\end{figure*}

\begin{figure*}[t]
\begin{center}
    \includegraphics[width=\linewidth]{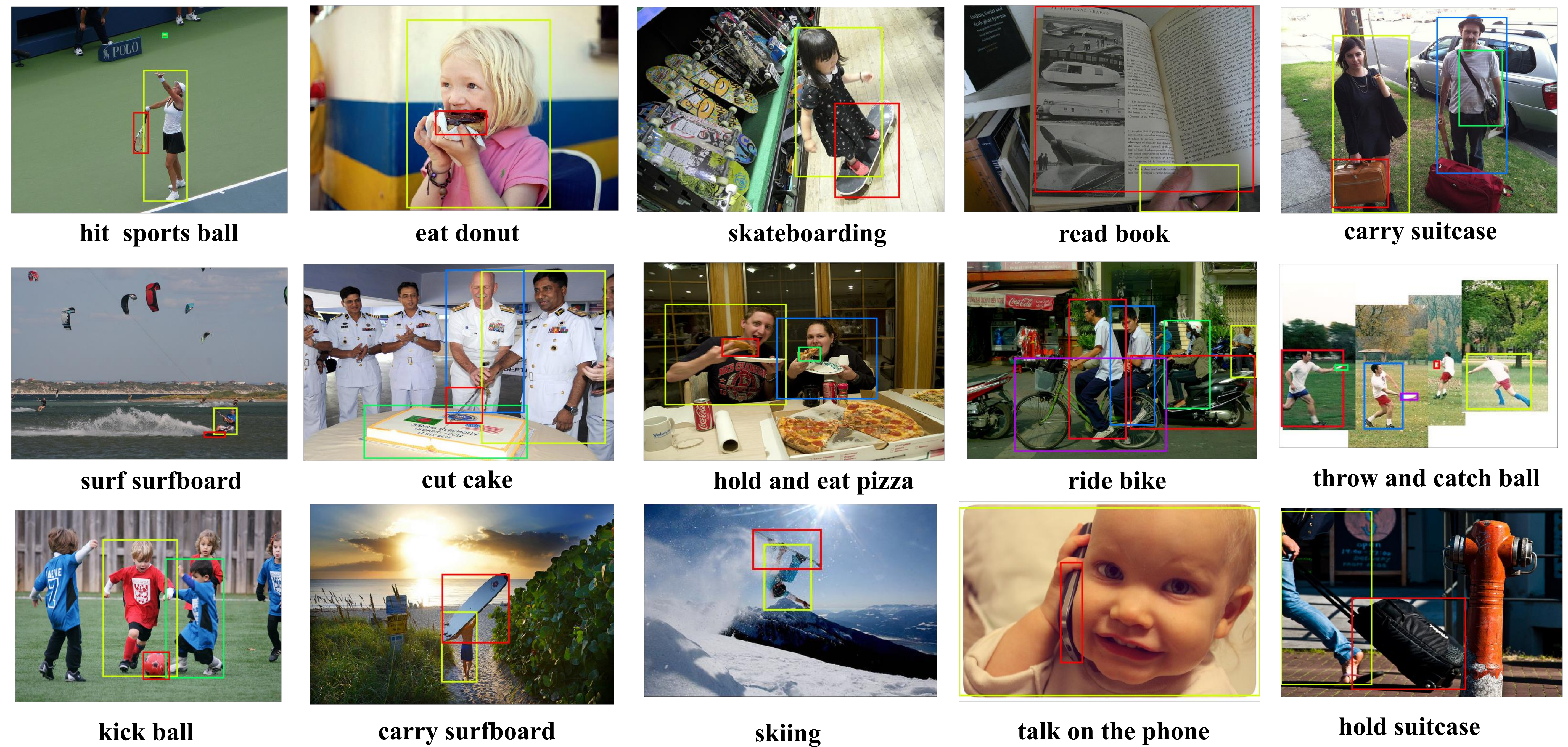}
   %\resizebox{0.8\textwidth}{!}{\includegraphics{S3D.png}}%
\end{center} \vspace{-0.3cm}
   \caption{Example detection results on V-COCO dataset. Each example can involve a single human- object interaction such as \textit{skateboarding} and \textit{eat donut }or multiple humans sharing the same interaction and object - \textit{hold} and \textit{eat pizza}, \textit{throw} and \textit{catch ball}.} \vspace{-0.1cm}
   \label{fig:V-COCO-qualitative}
\end{figure*}

We further evaluate the impact of contextual information on improving the classification capabilities of the multi-stream architecture. This is done by selecting different IoU thresholds in the range [0.1-0.9] used in the test evaluation of interaction recognition performance. Tab.~\ref{tab:over_thresh} shows the results on the V-COCO dataset. At a high threshold value (0.9), few ground-truth bounding-boxes are matched, whereas at a low threshold (0.1) most them are matched. Therefore, comparison at lower thresholds mainly focuses on the classification capabilities of our approach. Tab.~\ref{tab:over_thresh} shows that our approach is superior in terms of classification capabilities, compared to the baseline.

Tab.~\ref{tab:v-generalization} shows the generalization capabilities of our approach with respect to different network architectures. We perform experiments using VGG-16, ResNet-50 and ResNet-101, each pre-trained on the ImageNet dataset, as the underlying network architectures. In all cases, our approach provides consistent improvements over the baseline.

\begin{figure*}[t]
\begin{center}
    \includegraphics[width=\linewidth]{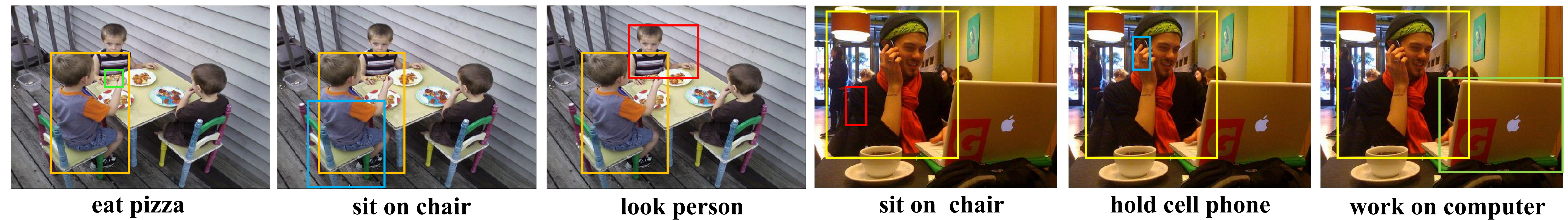}
  %\resizebox{0.8\textwidth}{!}{\includegraphics{S3D.png}}%
\end{center} \vspace{-0.3cm}
  \caption{Multiple interaction detection on V-COCO. Our approach detects human instance doing multiple (different) actions and interacting with various objects (represented with different colors). In all cases, the detected agent is represented with the same color.} \vspace{-0.3cm}
  \label{fig:V-COCO-camparsion}
\end{figure*}

\begin{figure*}[t]
\begin{center}
    \includegraphics[width=\linewidth]{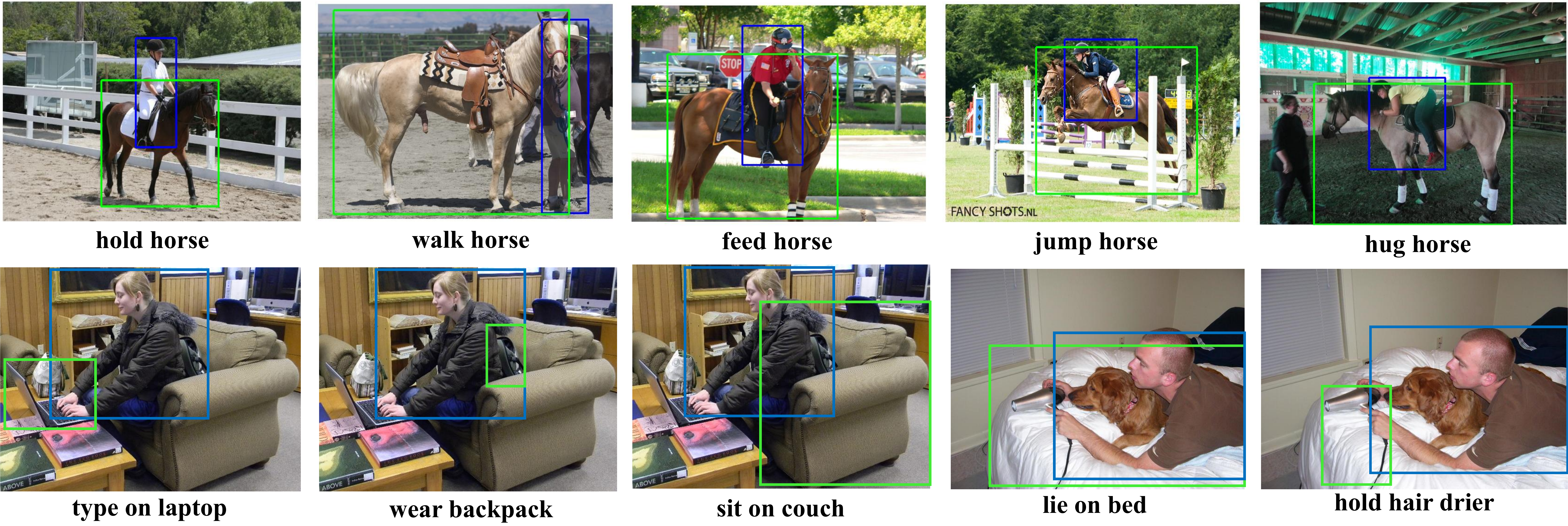}
   %\resizebox{0.8\textwidth}{!}{\includegraphics{S3D.png}}%
\end{center} \vspace{-0.3cm}
   \caption{Results on HICO-DET showing one detected triplet. Blue boxes represent a detected human instance, while the green boxes show the detected object of interaction. Our approach detects various fine-grained interactions (top row) and multiple interactions (second row).}\vspace{-0.3cm}
   \label{fig:HICO-qualitative}

\end{figure*}
\noindent \textbf{Comparison with State-of-the-art:} In Tab.~\ref{tab:v-coco-compare}, we compare our approach with state-of-the-art methods in the literature on the V-COCO dataset. Among existing works, InteractNet \cite{gkioxari2017interactnet} jointly learns to detect humans, objects and their interactions achieving a mAP$ _{role} $ of 40.0. The GPNN approach \cite{qi2018learning} integrates structural information in a graph neural network architecture and provides a mAP$ _{role} $ of 44.0. The iCAN approach \cite{gao2018ican} combines human, object and their pairwise interaction streams in an early fusion manner using the standard appearance features and bottom-up attention strategy. Our approach sets a new state-of-the-art on this dataset by achieving a mAP$ _{role} $ of 47.3.

\noindent \textbf{Qualitative Comparison:} Fig.~\ref{fig:compare_attention} shows comparison between the attention maps obtained using our approach and iCAN \cite{gao2018ican} on example images from the V-COCO dataset. Note that the attention maps in iCAN \cite{gao2018ican} are constructed using standard appearance features. In contrast, the attention maps in our approach are constructed using contextual appearance features generated using the context aggregation and local encoding blocks in our context-aware appearance module. Our attention maps focus on relevant regions in the human and object branches that are likely to contain human-object interactions (\eg, in case of \textit{throwing frisbee} and \textit{riding bike}). In addition, for both single and multiple human-object interactions, our approach produces more anchored attention maps compared to the iCAN method.

Fig.~\ref{fig:V-COCO-qualitative} shows examples showing both single human-object interactions such as \textit{skateboarding} and \textit{eat a donut}, and multiple humans sharing same interaction and object -- \textit{holding} and \textit{eating pizza}, \textit{throw} and \textit{catch ball}. Fig.~\ref{fig:V-COCO-camparsion} shows examples of a human performing multiple interactions.

\subsection{Results on HICO-DET and HCVRD datasets} %Results on V-COCO:
% \noindent \textbf{Comparison with State-of-the-art:} Following \cite{chao2018}, we report results on three different HOI category sets: full, rare, and non-rare with two different settings of Default and Known Objects. Tab.~\ref{tab:hico-det-compare} compares the performance of our approach with the existing methods. Our approach outperforms the state-of-the-art in all three category sets under both Default and Known Object settings. The relative gain of $9.4\%$, $6.7\%$, and $9.8\%$ is obtained over the best existing method on all three sets in Default settings. Fig.~\ref{fig:HICO-qualitative} shows example detections on HICO-DET.
On HICO-DET we report results on three different HOI category sets: full, rare, and non-rare with two different settings of Default and Known Objects \cite{chao2018}. Our approach outperforms the state-of-the-art in all three category sets under both Default and Known Object settings (see Tab.~\ref{tab:hico-det-compare}. The relative gain of $9.4\%$, $6.7\%$, and $9.8\%$ is obtained over the best existing method on all three sets in Default settings. Fig.~\ref{fig:HICO-qualitative} shows results on HICO-DET.
On HCVRD dataset, iCAN achieves top-1 and top-3 accuracies at R@50 of 33.8 and 48.9, respectively. Our approach outperforms iCAN with top-1 and top-3 accuracies at R@50 of 37.1 and 51.3, respectively. Similarly, our approach provides superior results at R@100 (top-3 accuracy of iCAN: 49.4 vs. top-3 accuracy of ours: 51.9).

\begin{table}[t]
\begin{center}
\resizebox{\linewidth}{!}{
\begin{tabular}{l|ccc |ccc}
\hline
 &\multicolumn{3}{c|}{Default}  &\multicolumn{3}{c}{ Known Object} \\
\cline{2-7}
%-------------------------------------------------------------------------
% &\multirow{2}{c}{method}
Methods & full & rare & non-rare &full & rare & non-rare\\
\hline
Shen \etal, \cite{shen18}  & 6.46 & 4.24 & 7.12 &- &- &- \\
Chao \etal, \cite{chao2018}  & 7.81 & 5.37 & 8.54 &10.41 &8.94 &10.85\\
InteractNet  \cite{gkioxari2017interactnet} & 9.94 & 7.16 & 10.77 &- &- &-\\
GPNN \cite{qi2018learning} & 13.11 & 9.34 & 14.23 &- &- &-\\
iCAN \cite{gao2018ican} & 14.84 & 10.45 & 16.15 & 16.43 & 12.01 &17.75\\ \hline
% Ours & \textbf{15.94}   &\textbf{10.86}   &\textbf{17.45} &\textbf{17.43} &\textbf{12.48} &\textbf{18.91}\\
Ours & \textbf{16.24}   &\textbf{11.16}   &\textbf{17.75} &\textbf{17.73} &\textbf{12.78} &\textbf{19.21}\\
\hline
\end{tabular}
}
\end{center}\vspace{-0.2cm}
\caption{State-of-the-art comparison on the HICO-DET using two different settings: Default and Known Object on all three sets (full, rare, non-rare). Note that Shen \etal \cite{shen18}, InteractNet  \cite{gkioxari2017interactnet} and GPNN \cite{qi2018learning} only report results on the Default settings. Our approach achieves a relative gain of $9.4\%$, $6.7\%$, and $9.9\%$ over the best existing method on all three HOI sets in Default settings.}\vspace{-0.4cm}\label{tab:hico-det-compare}
\end{table}

\section{Conclusion}
We propose a deep contextual attention framework for HOI detection. Our approach learns contextually-aware appearance features for human and object instances. To suppress the background noise, our attention module adaptively selects relevant instance-centric context information crucial for capturing human-object interactions. Experiments are performed on three HOI detection benchmarks: V-COCO, HICO-DET and HCVRD. Our approach has been shown to outperform state-of-the-art methods on all datasets. \\
\noindent\textbf{Acknowledgments:}
This work was supported by the National Natural Science Foundation of China (Grant~\# 61632018), Academy of Finland project number 313988 and the European Unions' Horizon 2020  (Grant~\# 780069).
%------------------------------------------------------------------------

%This work has been supported by the Academy of Finland project number 313988 \emph{Deep neural networks in scene graph generation for perception of visual multimedia semantics} and the European Union's Horizon 2020 research and innovation programme under grant agreement No 780069 \emph{Methods for Managing Audiovisual Data:Combining Automatic Efficiency with Human Accuracy}

{\small
\bibliographystyle{ieee_fullname}
\bibliography{egbib}
}

\end{document}